\documentclass[twoside,11pt]{article}

\usepackage{jmlr2e}
\usepackage{nccfoots}
\usepackage{tikz}
    \usetikzlibrary{arrows,mindmap,trees,shapes}
\usepackage{listings}
   \lstdefinestyle{opengm}{
      language=C++,
      basicstyle=\sffamily\scriptsize,
      columns=fullflexible,
      numbers=left,
      numberstyle=\sffamily\scriptsize,
      escapechar=^,
      xleftmargin=7ex
}
\usepackage[]{caption}
\usepackage{framed}
    \definecolor{shadecolor}{rgb}{0.9,0.9,0.9}
\usepackage[pdftex,colorlinks,linkcolor=black,citecolor=black,urlcolor=blue]{hyperref}

\long\def\symbolfootnote[#1]#2{\begingroup%
\def\thefootnote{\fnsymbol{footnote}}\footnote[#1]{#2}\endgroup}
\newcommand{\R}{\Omega}

\jmlrheading{?}{2012}{?-??}{??/??}{??/??}{Bjoern Andres, Thorsten Beier and Joerg H.~Kappes}

\ShortHeadings{OpenGM: A C++ Library for Discrete Graphical Models}{B.~Andres, T.~Beier, J.~H.~Kappes}
\firstpageno{1}

\begin{document}

\title{OpenGM: A C++ Library for Discrete Graphical Models}

\newcounter{customfootnote}

\author{%
\name Bjoern Andres\footnotemark[1] \email bandres@seas.harvard.edu \\
\name Thorsten Beier\footnotemark[1] \email thorsten.beier@iwr.uni-heidelberg.de \\
\name J\"org H.~Kappes\footnotemark[1] \email kappes@math.uni-heidelberg.de \\
\addr HCI, University of Heidelberg, Speyerer Str.~6, 69126 Heidelberg, Germany\\
\textnormal{\url{http://hci.iwr.uni-heidelberg.de/opengm2}}}
\editor{}

\maketitle

\begin{abstract}%   <- trailing '%' for backward compatibility of .sty file
OpenGM is a C++ template library for defining discrete graphical models and
performing inference on these models, using a wide range of state-of-the-art
algorithms.
No restrictions are imposed on the factor graph to allow for higher-order
factors and arbitrary neighborhood structures.
Large models with repetitive structure are handled efficiently because
(i) functions that occur repeatedly need to be stored only once, and (ii)
distinct functions can be implemented differently, using different encodings
alongside each other in the same model.
Several parametric functions (e.g.~metrics), sparse and dense value tables are
provided and so is an interface for custom C++ code.
Algorithms are separated by design from the representation of graphical models
and are easily exchangeable.
OpenGM, its algorithms, HDF5 file format and command line tools are modular and 
extendible.
\end{abstract}


\Footnotetext{*}{Authors contributed equally.}
\section{Introduction and Related Work}

Graphical models have become a standard tool in machine learning, and inference
(marginal and MAP estimation) is the central problem,
cf.~\cite{Nowozin-2011}.

These models can be defined rigorously as models of functions that factorize
w.r.t.~an associative and commutative operation,
cf.~\cite{werner-2008}.
The C++ library OpenGM is based on this general definition that allows for a
unified treatment of accumulative operations on such functions, including
optimization, summation (marginalization), conjunction and disjunction. It
provides a variety of inference algorithms\footnote{Not all algorithms can be 
used with each semi-ring.} beyond message passing
(Fig.~\ref{figure:algorithms}).
It can deal efficiently with large scale problems, since (i) functions that
occur repeatedly need to be stored only once and (ii) when functions require
different parametric or non-parametric encodings, multiple encodings can be used
alongside each other, in the same model. No restrictions are imposed on the
factor graph and the operations of the model, and the file format handles user
extensions automatically. Furthermore, OpenGM is a template library in which
elementary data types can be chosen to maximize efficiency.

Existing libraries do not have all of these properties.
MRF-lib \citep{szeliski-2008} is restricted to the min-sum semi-ring and
second-order grid graphs. While it is highly efficient on these models, it is
also specialized to these and not easily extendible.
In contrast, libDAI~\citep{mooij-2010} supports max-product and sum-product 
semi-rings which are hard-coded. The main drawback of libDAI is that it supports 
only dense value tables to encode functions which becomes prohibitive for models 
with many labels and higher-order factors.
Similar to libDAI, FastInf \citep{jaimovich-2010} focuses on message passing and
does not impose any restrictions on the factor graph. In contrast to libDAI, it
supports shared functions and different function types in a so-called
relational model that is similar in spirit to the design of OpenGM. However,
FastInf supports only sum-product semi-rings and, unlike OpenGM, has no generic
template abstraction of semi-rings.
The recently published library grante
\citep{grante}
provides shared functions and different function types. Furthermore, it comes
with a set of learning methods. Unlike OpenGM, it is not template based, limited
in its inference methods and published under a proprietary license.

The generality of OpenGM comes at the cost of performance. And yet, OpenGM is 
only slightly slower than libDAI when running loopy belief propagation on a grid 
graph. The highly optimized code of MRF-LIB is twice as fast for general 
second-order factors and 20 times as fast for standard metrics.

OpenGM is modular and extendible. The graphical model data structure, inference
algorithms and different encodings of functions interoperate through
well-defined interfaces.

\section{Mathematical Foundation}
\label{section:math}

\begin{figure}
\centering
\small
\tikzstyle{algorithm} = [rectangle, rounded corners, minimum width=19ex, minimum height=4ex, draw=green, fill=green!10]
\begin{tikzpicture}[xscale=3, yscale=-0.7]
\node [] at (0, 0.1) {Message passing};
\node [algorithm] at (0, 1) {Loopy BP};
\node [algorithm] at (0, 2) {TRBP};
\node [algorithm] at (0, 3) {TRW-S};

\node [] at (1, 0.1) {Graph cut};
\node [algorithm] at (1, 1) {$\alpha$-expansion};
\node [algorithm] at (1, 2) {$\alpha\beta$-swap};
\node [algorithm] at (1, 3) {QPBO};

\node [] at (2, 0.1) {Search};
\node [algorithm] at (2, 1) {ICM};
\node [algorithm] at (2, 2) {LazyFlipper};
\node [algorithm] at (2, 3) {LOC};

\node [] at (3, 0.1) {Sampling};
\node [algorithm] at (3, 1) {Gibbs};
\node [algorithm] at (3, 2) {Swendsen-Wang};

\node [] at (4, 0.1) {LP / ILP};
\node [algorithm] at (4, 1) {\scriptsize Dual decomposition};
\node [algorithm] at (4, 2) {Branch \& cut};
\node [algorithm] at (4, 3) {A$^{*}$};

\end{tikzpicture}
\caption{\small Algorithms provided by OpenGM:
Loopy BP \citep{pearl-1988,kschischang-2001},
TRBP \citep{wainwright-2008},
TRW-S \citep{kolmogorov-2006},
$\alpha$-expansion, $\alpha\beta$-swap \citep{boykov-2001},
QPBO \citep{rother-2007},
ICM \citep{besag-1986},                                             
Lazy Flipper \citep{andres-2010},
LOC \citep{jung-2009},
Swendsen-Wang sampling \citep{barbu-2005},
Dualdecomposition (subgradient and bundle-methods) \citep{kappes-2012,komodakis-2010}, native LP and
Branch \& cut using IBM ILOG Cplex,
A$^{*}$ \citep{bergtholdt-2010}. }
\label{figure:algorithms}
\end{figure}

OpenGM is built on a rigorous definition of the syntax and semantics of a
graphical model.
The syntax determines a class of functions that factorize w.r.t.~an
associative and commutative operation. In a probabilistic model, it determines
the conditional independence assumptions. The semantics specify the operation
and one function out of the class of all function that are consistent with the
syntax.

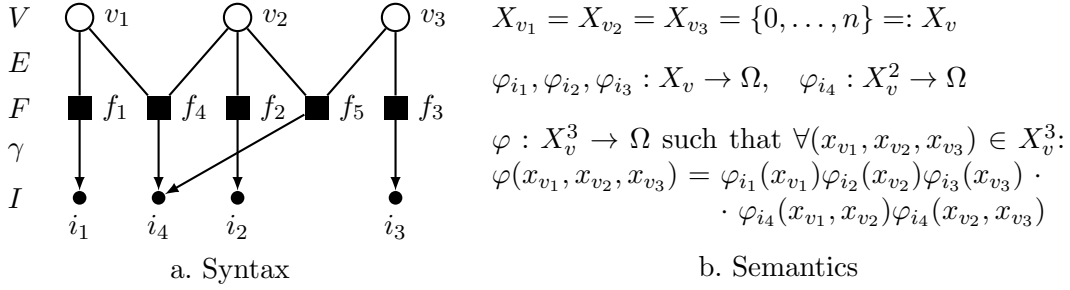
\begin{figure}
\begin{minipage}{0.44\textwidth}
\begin{tikzpicture}[xscale=0.7,yscale=-0.8,auto]
\tikzstyle{vnode}=[circle,draw=black,thick]
\tikzstyle{fnode}=[scale=1.2,fill=black]
\tikzstyle{gnode}=[circle,scale=0.5,fill=black]
\tikzstyle{edge}=[draw,thick]
\tikzstyle{dedge}=[draw,thick,-latex]
\foreach \place/\name/\word in {{(0,0)/v1/$v_1$}, {(3,0)/v2/$v_2$}, {(6,0)/v3/$v_3$}}
    \node[vnode] (\name) [label=right:\word] at \place {};
\foreach \place/\name/\word in {{(0,1.5)/f1/$f_1$}, {(1.5,1.5)/f4/$f_4$}, {(3.0,1.5)/f2/$f_2$}, {(4.5,1.5)/f5/$f_5$}, {(6,1.5)/f3/$f_3$}}
    \node[fnode] (\name) [label=right:\word] at \place {};
\foreach \place/\name/\word in {{(0,3)/g1/$i_1$}, {(1.5,3)/g4/$i_4$}, {(3,3)/g2/$i_2$}, {(6,3)/g3/$i_3$}}
    \node[gnode] (\name) [label=below:\word] at \place {};
\path[edge] (v1) -- (f1);
\path[edge] (v1) -- (f4);
\path[edge] (v2) -- (f2);
\path[edge] (v2) -- (f4);
\path[edge] (v2) -- (f5);
\path[edge] (v3) -- (f3);
\path[edge] (v3) -- (f5);
\path[dedge] (f1) -- (g1);
\path[dedge] (f2) -- (g2);
\path[dedge] (f3) -- (g3);
\path[dedge] (f4) -- (g4);
\path[dedge] (f5) -- (g4);
\node (V) [label=right:$V$] at (-1.75,0) {};
\node (E) [label=right:$E$] at (-1.75,0.75) {};
\node (F) [label=right:$F$] at (-1.75,1.5) {};
\node (gamma) [label=right:$\gamma$] at (-1.75,2.25) {};
\node (I) [label=right:$I$] at (-1.75,3) {};
\end{tikzpicture}\vspace{-4mm}
\begin{center}a.~Syntax\end{center}
\end{minipage}
\begin{minipage}{0.5\textwidth}
$X_{v_1} = X_{v_2} = X_{v_3} = \{0,\ldots,n\} =: X_v$\\[2ex]
$\varphi_{i_1}, \varphi_{i_2}, \varphi_{i_3}: X_v \to \R,
\hspace{2ex} \varphi_{i_4}: X_v^2 \to \R$\\[2ex]
$\varphi: X_v^3 \to \R$ such that $\forall (x_{v_1},x_{v_2},x_{v_3}) \in X_v^3$:
\begin{tabular}{@{}l@{\ }l@{\ }l}
$\varphi(x_{v_1},x_{v_2},x_{v_3})$ & $=$ & $\varphi_{i_1}(x_{v_1}) \varphi_{i_2}(x_{v_2}) \varphi_{i_3}(x_{v_3})\ \cdot$\\
& & $\cdot\ \varphi_{i_4}(x_{v_1}, x_{v_2}) \varphi_{i_4}(x_{v_2}, x_{v_3})$
\end{tabular}
\begin{center}b.~Semantics\end{center}
\end{minipage}
\caption{A factor graph $(V,F,E)$ describes how a function $\varphi$ decomposes
into a product of functions. In OpenGM, we extend this syntax by a set $I$ of
\emph{function identifiers} and a mapping $\gamma: F \to I$ that assigns one
function identifier to each factor. In the above example, the factors $f_4$ and
$f_5$ are mapped to the same function identifier $i_4$, indicating that the
corresponding functions $\varphi_{i_4}(x_{v_1}, x_{v_2})$ and
$\varphi_{i_4}(x_{v_2}, x_{v_3})$ are identical.}
\label{figure:syntax-semantics}
% such that $\forall (x,x') \in X_v^2$:\\
% $\varphi_{i_4}(x, x') = \left\{\begin{array}{ll}
%     0 & \textnormal{if}\ x = x'\\
%     r \in \R & \textnormal{otherwise}
% \end{array}\right.$\\
\end{figure}

The \emph{syntax}
(Fig.~\ref{figure:syntax-semantics}a)
consists of
a factor graph, i.e.~a bipartite graph $(V,F,E)$,
a linear order $<$ in $V$,
a set $I$ whose elements are called \emph{function identifiers}, and
a mapping $\gamma: F \to I$ that assigns one function identifier to each factor
such that only factors that are connected to the same number of variables can
be mapped to the same function identifier.
%for all $f,f' \in F$,
%$\gamma(f) = \gamma(f')$ implies $\deg(f) = \deg(f')$
%with
%$\deg(f) := |\{v \in V | (v,f) \in E\}|$.

For any $v \in V$ and $f \in F$, the \emph{factor} $f$ is said to \emph{depend}
on the \emph{variable} $v$ iff $(v,f) \in E$. $\mathcal{N}(f)$ denotes the set
of all variables on which $f$ depends and
$(v^{(f)}_j)_{j \in \{1,\ldots,|\mathcal{N}(f)|\}}$ the sequence of these
variables in ascending order.
Similarly, $(v_j)_{j \in \{1,\ldots,|V|\}}$ denotes the sequence of \emph{all}
variables in ascending order.

\emph{Semantics}
(Fig.~\ref{figure:syntax-semantics}b)
w.r.t.~a given syntax consist of
one finite set $X_v \not= \emptyset$ for each $v \in V$,
a commutative monoid $(\R,\odot,1)$ and
for any $i \in I$ for which there exists an $f \in F$ with $\gamma(f)=i$,
one function\footnote{The existence of $\varphi$ implies
$\forall f,f' \in F : \gamma(f) = \gamma(f') \Rightarrow
\forall j \in \{1,\ldots,\deg(f)\}: X_{ v^{(f)}_j } = X_{ v^{(f')}_j }$.}
$\varphi_i: X_{ v^{(f)}_1 } \times \cdots \times X_{ v^{(f)}_{|\mathcal{N}(f)|} } \hspace{-1ex} \to \R$.

The function from
$X := X_{v_1} \times \cdots \times X_{v_{|V|}}$
to
$\R$
\emph{induced} by syntax and semantics is the function
$\varphi: X \to \R$
such that
$\forall (x_{v_1}, \ldots, x_{v_{|V|}}) \in X$:
\begin{equation}
\varphi(x_{v_1}, \ldots, x_{v_{|V|}})
:=
\bigodot_{f \in F}
\varphi_{\gamma(f)} \left(
   x_{ v^{(f)}_1 }, \ldots, x_{ v^{(f)}_{|\mathcal{N}(f)|} }
\right) \enspace .
\label{eq:model}
\end{equation}

W.l.o.g., OpenGM simplifies the syntax and semantics by substituting
$V = \{0,\ldots,|V|-1\}$, equipped with the natural order, $F = \{0,\ldots,|F|-1\}$,
$I = \{0,\ldots,|I|-1\}$ and for each $v \in V$, $X_v = \{0,\ldots,|X_v|-1\}$.
A graphical model is thus completely defined by the number of
variables $|V|$, the number of labels $|X_v|$ of each variable $v \in V$, the
edges $E$ of the factor graph, the number of functions $|I|$, the assignment of
functions to factors $\gamma$, the commutative monoid $(\R,\odot,1)$ and one
function $\varphi_i$ for each function identifier $i \in I$.

Given a graphical model and, instead of just the commutative monoid
$(\R,\odot,1)$, a commutative semi-ring $(\R,\odot,1,\oplus,0)$, the problem of
computing
\begin{equation}
\bigoplus_{x \in X}
\varphi(x)
\qquad \textnormal{i.e.} \qquad
\bigoplus_{x \in X}
\bigodot_{f \in F}
\varphi_{\gamma(f)} \left(
   x_{ v^{(f)}_1 }, \ldots, x_{ v^{(f)}_{|\mathcal{N}(f)|} }
\right)
\label{eq:problem}
\end{equation}
is a central problem in machine learning with instances in
optimization $(\mathbb{R},+,0,\min,\infty)$,
marginalization $(\mathbb{R}^+,\cdot,1,+,0)$
and constrained satisfaction $(\{0,1\},\wedge,1,\vee,0)$.

\section{Using and Extending OpenGM}

The first step when using OpenGM is to construct a \emph{label space} that
determines the number of variables and the number of labels of each variable.
The next step is to fix the data type of the domain $\R$, the operation
$\odot$ and the way functions are encoded, by choosing the parameters of the
\emph{graphical model} class template as in the example below. 

To define a \emph{function} such as $\varphi_{i_4}(y_1,y_2)$, one needs to
indicate how many labels $y_1$ and $y_2$ have and set the parameters of the
function or fill its value table.
Once a function has been added to the model, it can be connected to several
\emph{factors} and thus assigned to different sets of variables.
This procedure is always the same, regardless of the number and type of classes
used to encode functions. Details are described in the users' section of the
manual.

Algorithms for optimization and inference are classes in OpenGM. To run an
algorithm, one instantiates an object of the class, providing a model
and optional control parameters, and calls the member function \emph{infer}, 
either without parameters or with one parameter indicating a \emph{visitor} 
(see example). Visitors are a powerful tool for monitoring and
controlling algorithms by code injection. Once an algorithm has terminated, 
results such as optima and bounds can be obtained via member functions. 
Detailed instructions can be found in the users' section of the manual.

OpenGM provides interfaces for custom algorithms, custom parametric functions,
custom discrete spaces and custom semi-rings. These interfaces are described in
the developers' section of the manual.

\begin{shaded}
\begin{lstlisting}[style=opengm]
typedef SimpleDiscreteSpace<size_t, size_t> Space;
Space space(numberOfVariables, numberOfLabels);^\\[-1.5ex]^
typedef OPENGM_TYPELIST_2(ExplicitFunction<float>, PottsFunction<float>) Functions;
typedef GraphicalModel<float, Adder, Functions, Space> Model;
Model gm(space);^\\[-1.5ex]^
ExplicitFunction<float> f1(&numberOfLabels, &numberOfLabels + 1);
f1(0) = ...; f1(1) = ...; ...
Model::FunctionIdentifier fid1 = gm.addFunction(f1);
const size_t variableIndex = 0;
gm.addFactor(fid1, &variableIndex, &variableIndex + 1);^\\[-1.5ex]^
PottsFunction<float> f2(numberOfLabels, numberOfLabels, 0.0f, 0.3f);
Model::FunctionIdentifier fid2 = gm.addFunction(f2);
size_t variableIndices[] = {variableIndex, variableIndex + 1};
gm.addFactor(fid2, variableIndices, variableIndices + 2);^\\[-1.5ex]^
typedef BpUpdateRules<Model, Minimizer> UpdateRules;
typedef MessagePassing<Model, Minimizer, UpdateRules, MaxDistance> BeliefPropagation;
BeliefPropagation::Parameter parameter(maxNumberOfIterations, convergenceBound, damping);
BeliefPropagation bp(gm, parameter);
MessagePassingVerboseVisitor<BeliefPropagation> visitor;
bp.infer(visitor);
vector<size_t> labeling(numberOfVariables);
bp.arg(labeling);
\end{lstlisting}
\end{shaded}

\section{Conclusion}

OpenGM is a C++ library for finite graphical models that provides
state-of-the-art inference algorithms. It widens the range of
models representable in software by allowing for arbitrary factor graphs and
semi-rings and by handling models with repetitive structure efficiently. It
is fast enough for prototype development even in settings where performance is
paramount. The modularity and extendibility of OpenGM, its command line tools
and file format have the potential to stimulate an exchange of models and
algorithms.

\newpage
{\small%\bibliographystyle{natbib}
\bibliography{references}}

% \footnotetext[2]{%
% LBP \citep{pearl-1988,kschischang-2001},
% TRBP \citep{wainwright-2008},
% TRW-S \citep{kolmogorov-2006},
% Graph cut, $\alpha$-expansion, $\alpha\beta$-swap \citep{boykov-2001},
% Fusion moves \citep{lempitsky-2010},
% QPBO \citep{rother-2007},
% Subgradient descent \citep{komodakis-2010},
% Swendsen-Wang sampling \citep{swendsen-1987}
% ICM \citep{besag-1986},
% Lazy Flipper \citep{andres-2010}}

\end{document}